# Safe AI for health and beyond

## Monitoring to transform a health service

Mahed Abroshan, Michael Burkhart, Oscar Giles, Sam Greenbury, Zoe Kourtzi, Jack Roberts, Mihaela van der Schaar, Jannetta S Steyn, Alan Wilson, May Yong

## Abstract

Machine learning techniques are effective for building predictive models because they identify patterns in large datasets. Development of a model for complex real-life problems often stop at the point of publication, proof of concept or when made accessible through some mode of deployment. However, a model in the medical domain risks becoming obsolete as patient demographics, systems and clinical practices change. The maintenance and monitoring of predictive models' performance post-publication is crucial to enable their safe and effective long-term use. We will assess the infrastructure required to monitor the outputs of a machine learning algorithm, and present two scenarios with examples of monitoring and updates of models, firstly on a breast cancer prognosis model trained on public longitudinal data, and secondly on a neurodegenerative stratification algorithm that is currently being developed and tested in clinic.

## Introduction

Visualise the following workflow: There is a stream of patients presenting into a hospital, each patient case is rich in accompanying data – both health and 'social' – all accumulated in a large database. For each patient, machine learning algorithms offer augmented intelligence to the clinician, for example, by locating patients in clusters characterising either medical diagnosis or prognosis. Armed with information about outcomes of treatments for similar patients, the clinician designs a treatment plan. The outcome of the treatment is monitored and recorded.

New patient data and well as updated records is precious resource. In today's algorithm-led domains, what should our considerations be in order to make full use of this resource [18]?

Data changes over time and in space, illustrating two orthogonal modes of change that models must contend with. Such changes in data inputs are commonly referred to as *drift* [20] result from temporal changes (for example, weight or smoking habits of patient population changing over time) or behavioural changes (for example, differences in data collection and interpretation under multiple clinical settings). They can also result from geographical and socio-economic changes (for example. patients at different hospitals may be different in demographic terms.)

Machine learning algorithms are trained on data. Because of data drift, the performance of a model for complex real-life problems does not peak and remain fixed at the point of publication or proof of concept or even point of deployment. Updated patient records contain information that: 1) enriches existing datasets and 2) may reflect emerging patterns. If the first is the case, machine learning algorithms *could* be updated to increase robustness; if the second, algorithms *should* be adjusted to reflect new data distributions. Either way, this requires an ongoing assessment of the effectiveness of treatment plans and if this is being done for many patients, the system – a 'Learning Machine' – can continuously identify optimum treatment plans.

A Learning Machine's strongest benefit is that the system remains safe and becomes more effective with time because data inputs are monitored, and the model is updated with the changing statistical properties of its inputs.

Monitoring and updating is crucial to high-stake environments that demand transparency and accountability. A model developed using retrospective data in the medical domain risks becoming obsolete as patient demographics and clinical practices change. One study has provided some insights into the speed in which performance deteriorates, with deterioration occurring as early as a year after deployment [19]. Another study links the performance deterioration to drifts in data recording practices [20].

> "Clinical data is highly dependent on the landscape of clinical practice as well as underlying population demographics and comorbidities, all of which vary over time. The complete utility of a healthcare model can be nearly impossible to ascertain unless one accounts for the inevitable effect of temporal dataset drift."
>
> *Nestor et al. [20]*

There are major research challenges that underpin this development:

The first lies in the creation of effective machine learning models for medical diagnosis and prognosis. This is especially challenging in healthcare as the results for a patient (or patients in a cluster) depend on complex factors such as medical histories, co-morbidities and demographic-dependent aetiology. This area is one in which many health-related AI applications are based. However, in the healthcare diagnosis or prognosis prediction domain, much focus is based on creating an instance of a model, that is, a model that is trained on a static dataset. When model development is complete, the emphasis will eventually shift to how to build the tools, infrastructure and regulations needed to efficiently deploy innovations in ML in clinical settings [18], while considering behavioural and temporal shift in data as described above.

Another challenge is data management: it is non-trivial to handle large, complex, volumes of multi-modal time series data, potentially stored across different IT systems. Care must be taken that data from a single patient such as observations, lab tests and treatments are recorded using the correct vocabularies and linked to form a coherent timeline of the patient's journey. If datasets from two hospitals are linked, common vocabularies and a good understanding of both hospitals' operations and practices are necessary to ensure no information is lost. In this scenario, there is the additional challenge of maintaining patient privacy. Hospitals balance the benefits of linking data for richer datasets and releasing data to enable research, against the possibility of patient privacy loss.

There are also the difficulties in designing and recording workflow feedback loops to provide transparency and accountability that healthcare systems demand. It is difficult to build user-trust in a system whose recommendations following model updates may change over time. There is also difficulty in communicating information that accompanies a prediction, such as confidence and uncertainty.

In the rest of this paper, we develop and apply the idea of a Learning Machine. We will be presenting the systems design of an ML workflow in the context from a Learning Machine. We will briefly describe both the research analytics and engineering challenges to be addressed in order to implement the workflow. We will then present two use cases where we explore the relationship between data drift (both temporal drift and changes in clinical settings) and machine learning performance. We will discuss what this means for the next steps in Health and propose how these challenges might be applicable outside Health.

## 1. Architecture of a Learning Machine

We will now build on the outline in the previous section and formalise the notion of a Learning Machine, characterised by four key phases:

1. New data records are generated from patients through a set of transformations including data collection and wrangling, with generated summary statistics and tests to ensure the data is correct.
2. Analytics are performed through machine learning.
3. Results are provided in the form of augmented intelligence for a decision-maker who uses this information to design an intervention.
4. Runs an evaluation of consequences from this intervention.

This is followed by a fifth phase:

5. A feedback loop to update records.

It is the existence of new data that enables updating and rerunning of analytics in the next cycle. Updated augmented intelligence aims towards improved 'optimum' interventions over time, with this form of 'learning' providing the central concept of a Learning Machine. This abstract idea provides the basis for a wide range of applications but, in the first instance, illustrate its application in health.

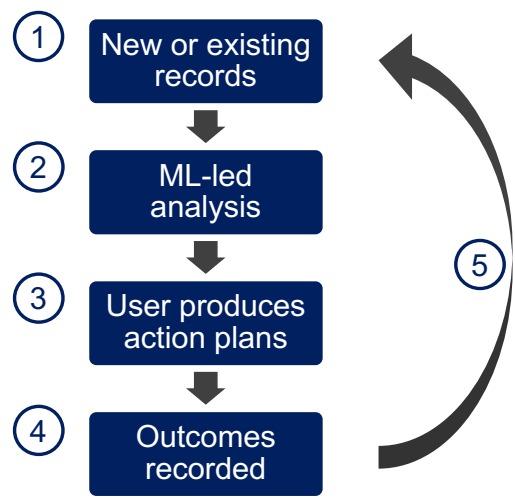

*Figure 1: An abstract depiction of the central workflow of a Learning Machine: 1) Add new or updated data records → Prepare data for analysis; 2) Perform ML-type analysis → Inference and predictions, explain analysis results; 3) Communicate to decision-makers → Specification of action plans; 4) Evaluate outcomes → update data records; 5) Feedback into system → New data generated. Step 1 repeats with updated records, and new samples.*

In Fig. 1, a Learning Machine is depicted in a generic healthcare system. We assume the core of the analysis of patient data is implemented through machine learning (ML) and the output offers augmented intelligence on diagnosis and prognosis to clinicians. They in turn produce treatment plans whose impacts are evaluated over time. These evaluations offer modified patient data for subsequent cycles.

To make this more tangible, consider the flow of patients into a hospital presenting with a heart problem. A machine learning algorithm will construct a diagnosis for an individual patient based on the patient's data and might locate the patient in a cluster, based on socio-economic factors and state of the cardiological disease. It may also offer a prediction given past disease trajectory: what is the probability of the patient having a heart attack in one year, two years and so on? These probabilities will be a function of the patient's medical history, including such complicating factors as co-morbidities', presenting a challenge for accurate prediction. The clinician, based on the augmented intelligence combined with professional judgement, will produce a treatment plan. For example, the plan might involve medications and informing the patient's GP of this predicted trajectory. This additional information enables the GP to better recommend lifestyle changes, to reduce risk and avert the predicted medical incidents.

There will be a variety of such plans from the population of clinicians, and these will be evaluated over time with the results fed back into the patients' data for later years. Since the new data will include the effectiveness of treatment plans, the machine will 'learn' and, over time, will deliver optimum and personalised treatment plans. This system is shown diagrammatically in Figure 2.

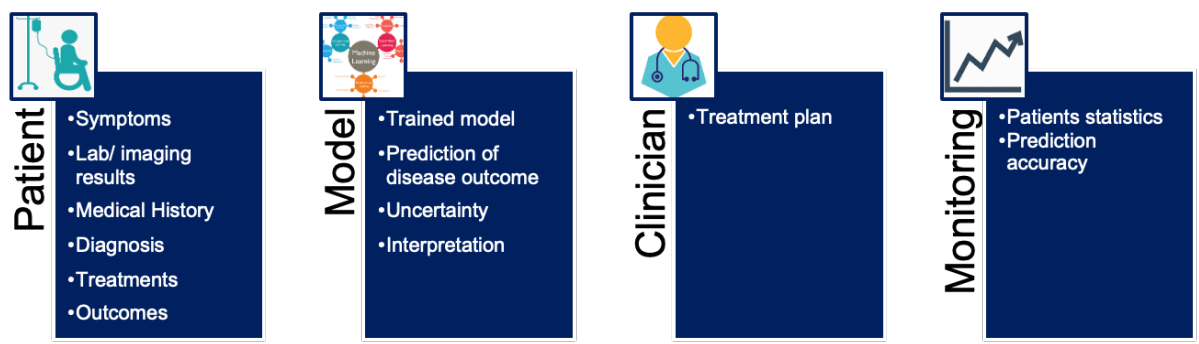

Figure 2: Depiction of an applied learning machine in the context of healthcare provision

## 2. Analytics challenges for a Learning Machine: How to detect drifts

This section of the learning machine infrastructure addresses how to detect differences resulting when data has shifted over time, or when the model is deployed on a different patient demographic.

In the healthcare domain, datasets typically contain many patients with their medical histories needing to be pieced together from multiple modalities and sources. This process demands considerable domain expertise, data linkage and wrangling, to avoid misinterpretation and minimise loss of information. As part of data wrangling, data quality reports and descriptive statistics should be generated as per drift monitoring below.

Drift monitoring means monitoring the properties of a dataset (for example, distributions of patient features and outcomes) and evaluate the performance of the model on the latest data. Machine learning engineers need to consider how a change of information content (for example, records of new lab tests, imaging modalities or genomics data) or granularity (for example data recording practices differ across clinics) affect machine learning decisions. New types of data (such as a new diagnostic test) may become available, which could increase the predictive power of the model if it were included. New screening procedures may introduce patients at earlier stages of diseases. Alternatively, model performance may drop over time due to improvements in treatment or other changes that are not reflected in the original model.

A package that can be used to track datasets for purpose of comparison measurement is the open-source Python package 'Learning Machines Drift' [23]. It works by allowing users to log a reference dataset as well as log subsequent new datasets. The package uses a set of statistical methods such as Kolmogorov-Smirnov and boundary adherence to quantify the differences between reference and subsequent data, and to assess if prediction outcomes and confidence values from the two datasets are drawn from different distributions. The package also uses machine learning approaches such as logistic detection and other classification methods to quantify differences, by measuring how well a model can be trained to classify between two datasets.

All these differences when measured will be flags to machine learning engineers about how the landscape of patient data is changed and offer them the opportunity to evaluate if the model is still appropriate.

## 3. Analytics challenges for a Learning Machine: When to retrain

This section of the Learning machine infrastructure addresses when to update machine learning models.

In the healthcare domain, machine learning algorithms are deployed for different purposes, with different requirements and work under different constraints. For example, when the goal is to suggest an optimal treatment plan for a given patient, a model that is designed to estimate the treatment effect and predict counterfactuals is needed [2,3]. When the goal is to predict which of the co-morbidities are more important for a particular individual's prognosis, a model capable of comparing competing risks is required [4]. When the goal is transparency, an example of a successful application of ML in healthcare is AutoPrognosis [12] which creates a priority list of patients with Cystic Fibrosis (CF), a disease with multiple co-morbidities [17] for lung transplant referral. It was shown using UK CF registry data that the ML method could improve the accuracy of clinical referral by 35% over traditional methods [13] while 'explaining' the important features leading to the ordering of the referral list.

However, after deployment and when data has drifted to effect performance deterioration of these algorithms, there is a need to retrain the model with new data or replace a model with one that works better with the distributions of the new data.

The machine will have to 'decide' when to retrain the model using information from drift monitoring, as described above. When this decision is taken, machine learning engineers must consider that a different machine learning approach may be more effective with coping with new data distributions. Therefore, the model retraining process should incorporate not only tuning the weights and hyperparameters of the previous model, but also the possibilities of 1) tuning hyperparameters using a different method or 2) using a completely different model or algorithm that yields better performance. This is a time-consuming task.

One option is to automate the comparison of a stable of approaches, as implemented in Auto-ML. In Auto-ML the goal is to select a method and tune a new model for a dataset without relying on a human expert. The authors use Bayesian optimization techniques [16] to find the right model and hyperparameters for a given dataset. When retraining is required, Auto-ML can retune AutoPrognosis and compare the updated model to other approaches in the Auto-ML stable, potentially suggesting a new model to be used. The Auto-ML approach for model selection would be beneficial in the context of a learning machine as they minimise the additional monitoring burden that arises.

## 4. Support infrastructure for a Learning Machine

In order to realise a Learning Machine architecture, there are several other important technical challenges we explore in this section. In reference to discussing these challenges, in Fig. 3 we provide a schematic of the components and the flow of data. Patients arrive in clinic and their data is transmitted through an API call to a deployment server where two events occur:

1. The deployment server contains a version of the current deployed model and provides a response containing suggested treatments and metrics to the clinician.
2. The data is transmitted to secure storage for further model development and monitoring.

Within secure storage the data is processed, monitored and incorporated into new proposed models. The silent monitoring processes include the use of descriptive statistics, performance monitoring and reproducibility checks. Periodically the model contained on the deployment server is updated with model proposals where it is deemed performance improvements can be made with the update. An additional data submission to the deployment server is necessary to capture outcomes of the patients.

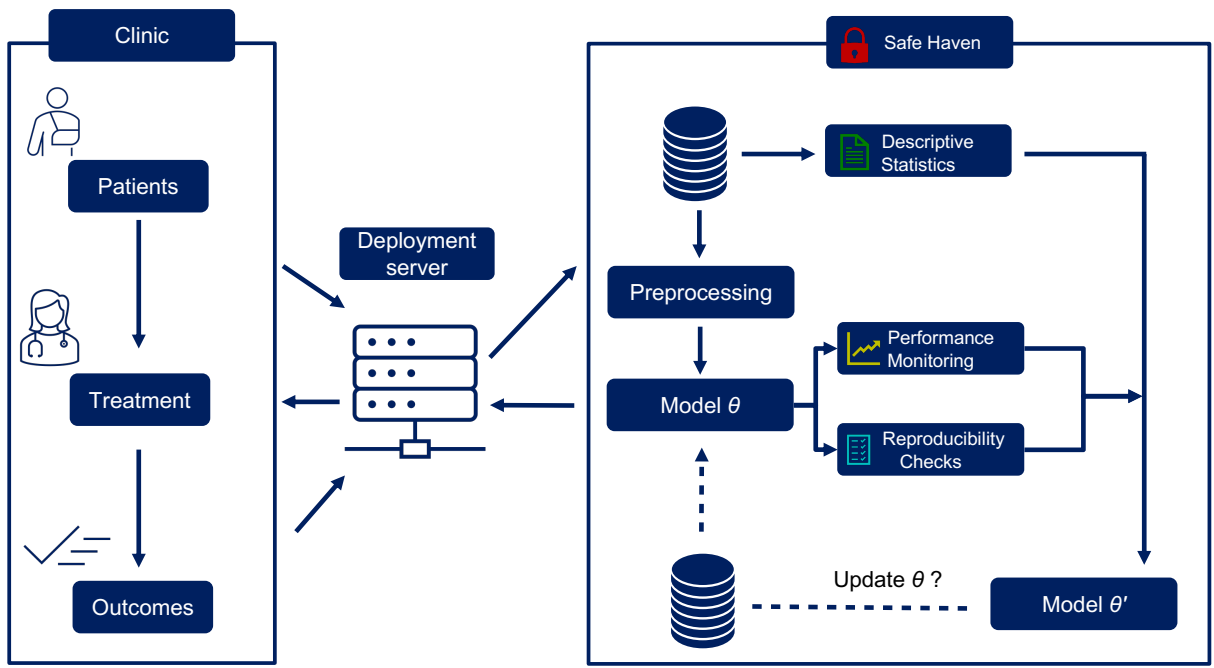

Figure 3: The components forming the support infrastructure of a learning machine deployed in the healthcare setting utilising secure storage and silent monitoring.

### Secure storage

Data will have to be securely stored and processed, keeping confidentiality and privacy in mind as the learning machine may use detailed patient-level and potentially (re-)identifiable records.

Environments such as the Data Safe Haven developed at The Alan Turing Institute [11] are designed to allow teams of researchers to analyse sensitive datasets safely and productively. It provides a framework for data owners to classify the sensitivity of their data, and automated tools to deploy infrastructure in the cloud tailored to that sensitivity tier. Examples of secure research environments in healthcare specifically include University Hospitals Birmingham's PIONEER [14].

### Continuous safety monitoring

Tests must be in place to show that the system will "do no harm"[21]. This is currently maintained in our recommended procedures by ensuring that the clinicians can combine the intelligence offered with their clinical judgement. As time progresses, it may be that more of the process can be automated.

### Transparency

The workings of the learning machine need to be transparent in order to generate trust and enable accountability. Transparency in this case relates to interpretability and uncertainty.

#### *Interpretability*

Since the learning machine will interact with human experts it is vital that predictions made by the model are interpretable. There are several notions of interpretability in the literature [5]. Instance-wise feature selection is one mode, where the 'interpreter' provides the most important features for making a particular prediction for each instance [6]. Another mode of interpretation is uncovering the statistical interactions captured by the model [7]. Depending on the task in hand, one of these methods could be more suitable and useful for human specialist.

#### *Uncertainty*

A model makes predictions; these predictions are paired with information denoting levels of uncertainty about the prediction. When the learning machine is used in sensitive applications like healthcare or criminal justice it is crucial that we provide uncertainty levels to human experts such that they can decide how to use the model output. Several different methods can be used for estimating uncertainty. Some models automatically provide uncertainty scores through their output being a continuous score that may be given a probabilistic interpretation representing the aleatoric uncertainty of individual outcomes. For instance, a neural network trained for classification usually gives the probability that the subject belongs to a particular class, the largest of these probability scores is an indicator for the uncertainty level. In addition, an understanding of the epistemic uncertainty associated with inferred model parameters and prediction outputs itself is important. Several methods are devised in the literature for estimating epistemic uncertainty of model outputs for a general model as a post processing step, which can be used in LM setting [8]. The uncertainty estimation can also be used to decide when to trigger model retraining. When the confidence intervals start to become larger, it is an indication that the distribution of the input has changed [9]. This issue has been studied in the literature about out-of-distribution (OOD) uncertainty [10].

## 5. Test cases

This section describes through examples how the concept of Learning Machines could be used to support two healthcare-related scenarios.

### Scenario 1: Breast cancer prognosis

The first scenario describes how we measure differences between earlier and latter years' breast cancer data and to examine if a survival prognosis model that has been built using earlier year's clinical data would still perform as well on cases from latter years.

The Surveillance, Epidemiology, and End Results (SEER) Program [22] has collected data on cancer diagnoses and outcomes in the USA since the 1970s. In total, the SEER datasets typically contain upwards of 5 million tumours across 10% of the US population, depending on the specific subset used. They include anonymised patient-level data, including demographics as well as information on the diagnosis (such as the type, stage, and grade of the tumour), treatments and survival outcome for each patient.

There are several changes we may expect to see from data over the last three decades. For example, detected tumour sizes may become smaller over time, due to changes in clinical practices such as increased screening. There are also demographic changes, for example smoking cessation or increased alcohol intake, which affects the age distribution of the patients. Depending on the features that the model relies on for making a prediction, these changes could be the types of drift that affects model performance. A learning machine will flag drift by continuous appraisal of inputs, model output and actual patient outcome. Without infrastructure to monitor data in place the machine learning model may become obsolete, and, without notice, performance may deteriorate and affect patient safety adversely.

We have used SEER to demonstrate the learning machines concept on a prognostic question: *given a breast cancer diagnosis, what is the 5-year survival probability for a patient?*

The first step we took is to visualise the differences in datasets across the years. Figure 4 below shows a screenshot from the tool that we developed to explore SEER data, select data for training models, and compare models' performances across time. In the top section, we have a timeline that shows the years from which data is available. We have checked the timeline to visualise data from years 1982-1992 and to use this to train a model. The section on descriptive statistics shows patient frequencies of survival rates, cancer stages and tumour grades and sizes, from the selected years.

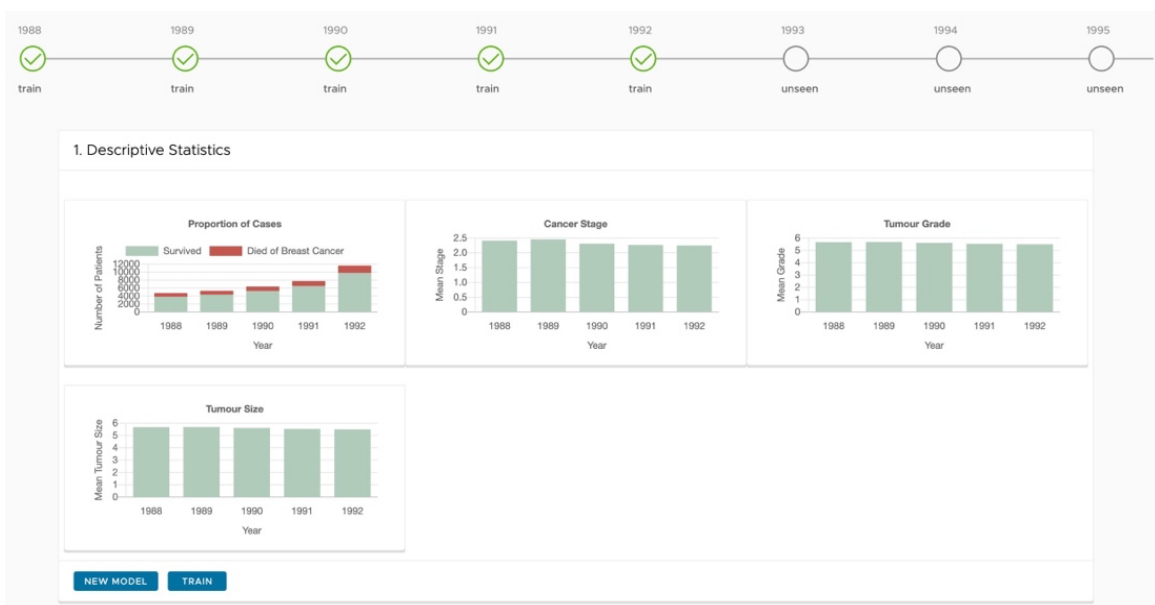

*Figure 4: This figure shows the frequency of survival rates, cancer stages, tumour grades and sizes from 1982 to 1992.*

Using the tool that we built, predictions from models trained on different datasets can be stored, so that model performance can be compared over time. For example. we can train models with data from a set of years and test them on data from subsequent years, to assess whether that model would have continued to accurately predict survival outcomes for patients diagnosed in the years ahead. With advances in cancer treatments, we may expect the original model to underestimate survival chances in later years.

Figure 5 below shows predictions, survival outcome and feature importance from predictions of three models, for two patients. Figure 5a represents a set of predictions for a patient who died of breast cancer within 49 months; models trained on data from different years erroneously makes confident predictions for survival, though with mortality rate increasing with each model iteration. Figure 5b shows a set of predictions for patient survival where mortality rate remains consistent over time. The final row of in Figure 5a and 5b shows the feature importance for predictions from each model allowing visualisation of drift in model parameters.

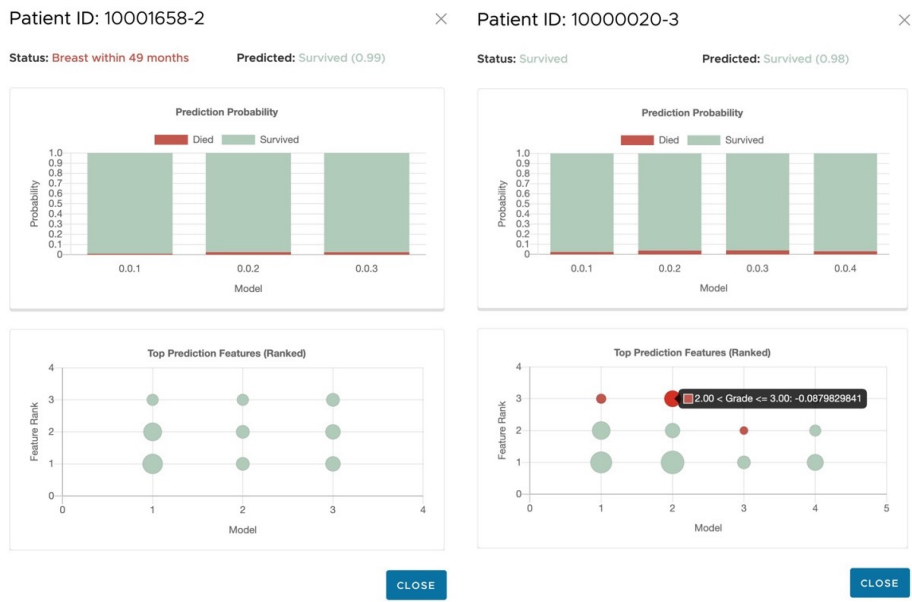

*Figure 5: Predictions from three models for two patients. The bar charts show respective patient probability of survival as produced by three models trained on different years; the bubble charts show feature importance of inputs from the models.*

## Conclusion

Our experiments showed that model performance did change with new data, but not always as expected. For example, older models sometimes outperformed models trained on more recent patients or models trained on larger datasets.

This demonstrates the importance of having infrastructure to continually monitor the properties of the dataset and appraise the performance of the model on contemporary patients as part of the learning machine.

This also demonstrates that the naïve approach of simply retraining models with the latest datasets, or with an amalgamation of new and old data, is no guarantee of an updated model with better performance.

## Scenario 2: Silent monitoring for hospital-deployed early detection of Alzheimer's algorithm

A digital tool is currently being developed for integration at the Cambridge University Hospital (CUH) Memory Clinic to distinguish mild cognitive impairment (MCI) from Alzheimer's disease. The machine learning algorithm was trained on the data from Alzheimer's Disease Neuroimaging Initiative (ADNI) and is being tested retrospectively on patients participating in the QMIN-MC (Quantitative MRI in NHS Memory Clinics) study. The goal is to deploy the method in memory clinics as a decision support tool for clinical diagnosis and prognosis.

The purpose of this example is to examine if a model trained on a research dataset from the United States, would perform as well when tested on real-world clinical data from memory clinics in the United Kingdom. This is an important question because patient outcome (both progressive and stable) will not be observable for many years. This means that we may not detect if the method performance begins to deteriorate.

The task of comparing model performance when deployed in different settings involves the following:

**Checks on the incoming data** (patient characteristics): Are new patients coming from a similar distribution as the training data? This can apply to patient demographics, cognitive scores and MRI scans, for example.

**Checks on the model outputs**: Are the distributions of the model outputs consistent with the training/historical data? For example, are the ratios of progressive to stable patients similar? Is the distribution of projection values similar?

**Checks on model performance**: Ensure the overall performance of the model (for example, the accuracy of the stable versus progressive prediction) is like performance with training data. Note the ability to do this is limited by the time taken to get ground truth labels – the time for a patient to be diagnosed as stable or progressive may be several years.

A core challenge is defining which metrics should be used to assess the model's current state, how this can be done in a timely manner, and what interventions can be taken when encountering unexpected behaviour.

*Initial findings and conclusion*
We conducted an initial proof-of-concept prediction task for distinguishing between MCI and Alzheimer's disease for patients from three different cohorts: ADNI (research cohort), NHS Memory Clinics (QMIN-MC) and MACC (Memory Aging and Cognition Centre memory clinic research cohort).

For the initial work, we focussed on MCI versus Alzheimer's disease, due to the availability of ground truth labels across datasets from the three different cohorts. This allows us to measure performance differences between the three sites as well as the presence of data drift and explore the potential for optimizing performance by combining data between sites for the trained model.

Our initial findings demonstrate that the deployment of an AI method to a new demography for this task is complex for optimisation, with trade-offs in dataset size and drift in features of importance for the classification task being important considerations. A learning machine that can provide a clear framework for monitoring decisions around which data to use in training is therefore vital.

*Learning machines as generic systems: wider areas of application.*
We have illustrated the concept through potential applications in medicine but as noted at the outset, it is essentially generic and is potentially widely applicable.  Consider the justice settings, where machine learning algorithms may provide predictions of violent behaviours or probability of an individual re-offending after release from prison. The patient from the healthcare examples becomes the prisoner in the justice context, and rather than a medical diagnosis we may use features relating to an individual's crime and behaviour whilst in prison. The learning machines infrastructure around this can remain the same. In fact, it may even become more poignant as the ethical considerations around algorithm-supported decision making in a justice context mean continual monitoring of its fairness and accuracy is essential and a legal requirement [15].

It may be possible to utilise learning machines in even more diverse settings – for example, in urban planning where the equivalent of medical histories are events in the life journeys of individuals and households. This would then become a tool for analysing some of the challenges of individual lives as a basis for informing future planning decisions such as transport or housing investment.

## 6. Concluding comments.

We have progressed through the proof-of-concept stage by building the software for a working 'learning machine'. We have illustrated its potential through two examples. The next steps will be to use more data sets and work towards collaborations with clinicians and decision makers utilising algorithms, models and evolving data over time in optimal ways.

## 7. Acknowledgement

This work includes data from the Quantitative MRI in NHS Memory Clinics (QMIN MC, [rittman.uk/qminmc](rittman.uk/qminmc)) study, supported by at the NIHR Cambridge Biomedical Research Centre (BRC-1215-20014); the views expressed are those of the author(s) and not necessarily those of the NIHR or the Department of Health and Social Care.

This work also includes data from MACC (Memory Aging and Cognition Centre memory clinic research cohort) funded by the National Medical Research Council of Singapore.

## Annex: System Diagram

Figure 4 conceptually shows the interaction between the components of the learning machine, patients, clinicians, and data scientists. The feedback referred to earlier is shown as an outer loop in this diagram. The inner loop is designed to evaluate model performance as the data changes over time and to signal the time point when the ML-model must be retrained. Table 1 below defines the labelled variables.

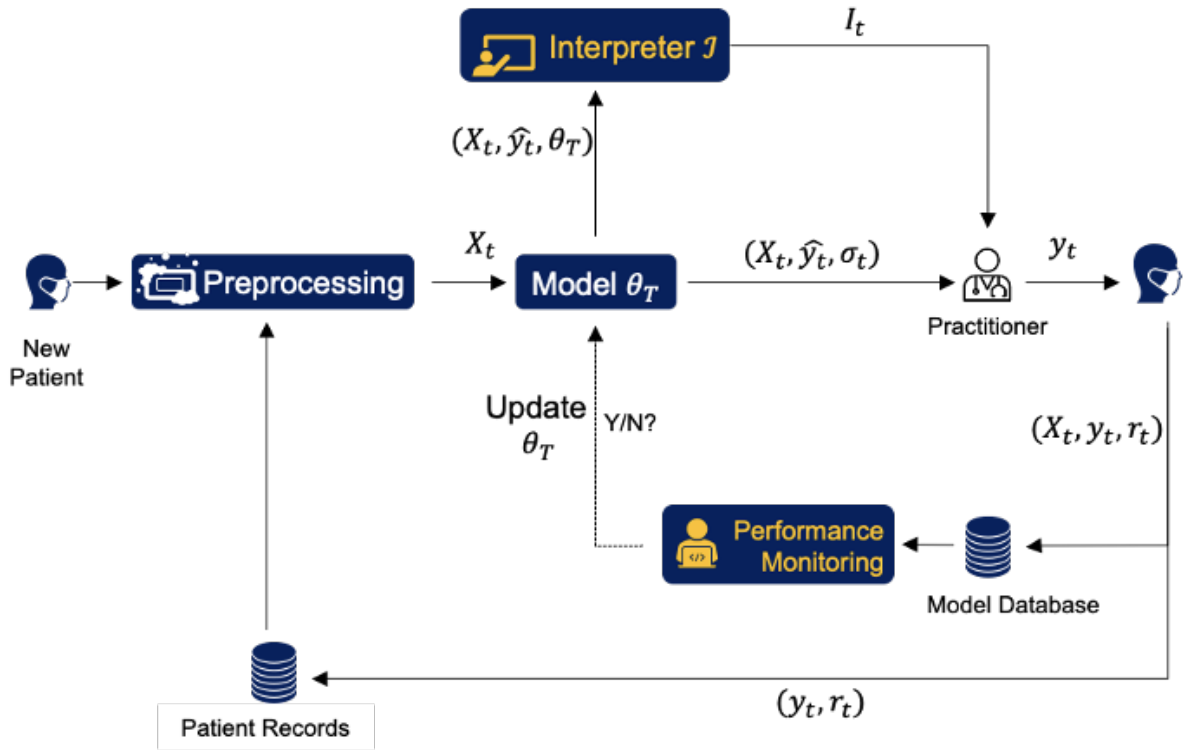

*Figure 6: Learning machine system diagram with symbols.*

| Variable | Meaning |
|---|---|
| $t$ | Current learning machine time step |
| $T$ | Time step when model was last retrained |
| $x_t$ | New patient data at time $t$ |
| $X_t$ | Patient data and history up to and including time $t$ |
| $\theta_T$ | Model trained using data up to time $T \leq t$ |
| $\hat{y}_t$ | Model prediction for patient arriving at time $t$ |
| $\sigma_t$ | Uncertainty of $\hat{y}_t$ |
| $I_t$ | Interpretation of $\hat{y}_t$ |
| $y_t$ | Practitioner's decision at time $t$ |
| $r_t$ | Treatment outcome for patient that arrived at time $t$ |

*Table 1: Definition of variables*

# References

[1] Nestor, B., McDermott, M., Boag, W., Berner, G., Naumann, T., Hughes, M.C., Goldenberg, A. and Ghassemi, M., 2019. Feature robustness in non-stationary health records: caveats to deployable model performance in common clinical machine learning tasks. arXiv preprint arXiv:1908.00690.

[2] Yoon, J., Jordon, J. and van der Schaar, M., 2018, February. GANITE: Estimation of individualized treatment effects using generative adversarial nets. In International Conference on Learning Representations.

[3] Bica, I., Alaa, A.M., Jordon, J. and van der Schaar, M., 2020. Estimating counterfactual treatment outcomes over time through adversarially balanced representations. arXiv preprint arXiv:2002.04083.

[4] Lee, C., Zame, W.R., Yoon, J. and van der Schaar, M., 2018, April. DeepHit: A Deep Learning Approach to Survival Analysis With Competing Risks. In AAAI (pp. 2314-2321).

[5] Lundberg, S.M. and Lee, S.I., 2017. A unified approach to interpreting model predictions. In Advances in neural information processing systems (pp. 4765-4774).

[6] Yoon, J., Jordon, J. and van der Schaar, M., 2018, September. INVASE: Instance-wise variable selection using neural networks. In International Conference on Learning Representations.

[7] Alaa, A.M. and van der Schaar, M., 2019. Demystifying Black-box Models with Symbolic Metamodels. In Advances in Neural Information Processing Systems (pp. 11304-11314).

[8] Alaa, A.M. and van der Schaar, M., 2020. Discriminative jackknife: Quantifying uncertainty in deep learning via higher-order influence functions. arXiv preprint arXiv:2007.13481.

[9] Lakshminarayanan, B., Pritzel, A. and Blundell, C., 2017. Simple and scalable predictive uncertainty estimation using deep ensembles. In Advances in neural information processing systems (pp. 6402-6413).

[10] Ovadia, Y., Fertig, E., Ren, J., Nado, Z., Sculley, D., Nowozin, S., Dillon, J., Lakshminarayanan, B. and Snoek, J., 2019. Can you trust your model's uncertainty? Evaluating predictive uncertainty under dataset shift. In Advances in Neural Information Processing Systems (pp. 13991-14002).

[11] Arenas, D., Atkins, J., Austin, C., Beavan, D., et al., 2019, Design choices for productive, secure, data-intensive research at scale in the cloud. arXiv preprint arXiv:1908.08737

[12] Alaa, A.M. and Schaar, M., 2018, July. AutoPrognosis: Automated clinical prognostic modeling via Bayesian optimization with structured kernel learning. In International conference on machine learning (pp. 139-148). PMLR.

[13] Alaa, A.M. and van der Schaar, M., 2018. Prognostication and risk factors for cystic fibrosis via automated machine learning. *Scientific reports*, *8*(1), pp.1-19.

[14] PIONEER Health Data Research Hub, University Hospitals Birmingham, https://www.pioneerdatahub.co.uk/

[15] Cabinet Office, 2021, Ethics, Transparency and Accountability Framework for Automated Decision-Making.

[16] Jones, D. R., Schonlau, M. and Welch, W. J. Efficient global optimization of expensive blackbox functions. Journal of Global Optimization 13, 455–492 (1998).

[17] Abroshan, M., Alaa, A.M., Rayner, O. and van der Schaar, M., 2020. Opportunities for machine learning to transform care for people with cystic fibrosis. *Journal of Cystic Fibrosis*, *19*(1), pp.6-8.

[18] Zhang, A., Xing, L., Zou, J. *et al.* Shifting machine learning for healthcare from development to deployment and from models to data. *Nat. Biomed. Eng* **6**, 1330–1345 (2022). https://doi.org/10.1038/s41551-022-00898-y

[19] Davis, S. E., Lasko, T. A., Chen, G., Siew, E. D. & Matheny, M. E. Calibration drift in regression and machine learning models for acute kidney injury. *J. Am. Med. Inform. Assoc.* **24**, 1052–1061 (2017).

[20] Nestor, B., McDermott, M. B. A. & Boag, W. Feature robustness in non-stationary health records: caveats to deployable model performance in common clinical machine learning tasks. Preprint at https://doi.org/10.48550/arXiv.1908.00690 (2019)

[21] Wiens, J., Saria, S., Sendak, M., Ghassemi, M., Liu, V. X., Doshi-Velez, F., … & Goldenberg, A. (2019). Do no harm: a roadmap for responsible machine learning for health care. *Nature medicine*, *25*(9), 1337-1340. https://doi.org/10.1038/s41591-019-0548-6

[22] Surveillance, Epidemiology, and End Results (SEER) Program (www.seer.cancer.gov) SEER*Stat Database: Incidence - SEER Research Data, 8 Registries, Nov 2021 Sub (1975-2019)

[23] Alan Turing Institute, Learning Machines Drift, (2023), GitHub repository, https://github.com/alan-turing-institute/learning-machines-drift